\documentclass{article}
\usepackage{collas2023_conference,times}
\usepackage{easyReview}
\usepackage{mathtools, amsmath, amssymb, amsfonts, bm, nicefrac, dsfont}
\usepackage{enumitem}
\usepackage{booktabs}
\usepackage{makecell}

\usepackage{hyperref}
\hypersetup{
    colorlinks=true,
    linkcolor=red,
    filecolor=magenta,
    urlcolor=blue,
    citecolor=purple,
    pdftitle={Overleaf Example},
    pdfpagemode=FullScreen,
    }

\title{Substituting Data Annotation with Balanced Updates and Collective Loss in Multi-label Text Classification \thanks{\textit{\underline{Citation}}: Ozmen, M. Cotnareanu, J. and Coates, M. Substituting Data Annotation with Balanced Updates and Collective Loss in Multi-label Text Classification. \textit{Proc. Conf. Lifelong Learning Agents (CoLLAs)}, 2023. \\ 
The implementation is available at \url{https://github.com/muberraozmen/BNCL}}}

\author{Muberra Ozmen \\
McGill University \\
Montreal, Canada \\
\texttt{muberra.ozmen@mail.mcgill.ca} \\
\And
Joseph Cotnareanu \\
McGill University \\
Montreal, Canada \\
\texttt{joseph.cotnareanu@mail.mcgill.ca} \\
\And
Mark Coates \\
McGill University \\
Montreal, Canada \\
\texttt{mark.coates@mcgill.ca} \\
}

\collasfinalcopy 

\begin{document}
\maketitle
\begin{abstract}
    Multi-label text classification (MLTC) is the task of assigning multiple labels to a given text, and has a wide range of application domains. Most existing approaches require an enormous amount of annotated data to learn a classifier and/or a set of well-defined constraints on the label space structure, such as hierarchical relations which may be complicated to provide as the number of labels increases.  In this paper, we study the MLTC problem in annotation-free and scarce-annotation settings in which the magnitude of available supervision signals is linear to the number of labels.  Our method follows three steps, (1) mapping input text into a set of preliminary label likelihoods by natural language inference using a pre-trained language model, (2) calculating a signed label dependency graph by label descriptions, and (3) updating the preliminary label likelihoods with message passing along the label dependency graph, driven with a collective loss function that injects the information of expected label frequency and average multi-label cardinality of predictions. The experiments show that the proposed framework achieves effective performance under low supervision settings with almost imperceptible computational and memory overheads added to the usage of pre-trained language model outperforming its initial performance by 70\% in terms of example-based F1 score.        
\end{abstract}
\section{Introduction}
Multi-label text classification (MLTC) is the task of selecting the correct subset of labels for each text sample in a corpus. MLTC has numerous applications, such as tagging articles with the most relevant labels or recommending related search engine queries \citep{mlcReview, xmlReview}. The majority of the literature \citep{XML-CNN, MLRNN, AttentionXML, MrMP} addresses the MLTC problem in a supervised setting, relying upon an abundance of annotated data. Despite their impressive classification performance on benchmark research datasets, most of these methods remain inapplicable in real-world applications due to the high cost of annotation. 

More recently, there has been an increasing focus on the single-label text classification problem with less \citep{SemiTC} or no \citep{LOTClass} annotated data. The adaptation of methods to the multi-label scenario, however, is not straightforward and often results in significant performance deterioration. One exceptional study by \cite{TaxoClass} considers hierarchical multi-label text classification without annotated data, but the algorithm requires a strict label taxonomy. Such extensive, and restrictive, prior information on the label space structure is not generally available.

In this work, we study generalized multi-label text classification, focusing on the limited annotated data setting while avoiding assumptions about the availability of strong structural information. We use a pre-trained language model to obtain preliminary label predictions using a natural language inference (NLI) framework. Pre-trained language models are trained on large-scale corpora which makes them better at recognizing patterns and relationships in natural language and allows them to handle rare words and phrases that may not appear frequently in a specific training dataset. We develop a framework that incorporates label dependencies and easily obtained supervision signals to adapt the predictions made by the pre-trained language model to the contextual properties of the specific data under study. Our experiments show that the proposed framework is efficient and effective in terms of improving the prediction performance. In summary, our key contributions are:
\begin{enumerate}[leftmargin=*]
    \item We develop a framework for multi-label text classification in two limited supervision settings:
    \begin{itemize}[leftmargin=*]
        \item[-] (1) label descriptions and (2) expected label observation probabilities and average subset cardinality or,
        \item[-] (1)  label descriptions and (2) a small set of annotated data.
    \end{itemize}
    \item We use multiple external linguistic knowledge bases: (1) a pre-trained language model that provides preliminary label likelihoods; (2) a set of pre-trained word embeddings to calculate signed label dependency graph.
    \item We propose a model that updates preliminary likelihoods by modelling label dependencies based on balance theory and by effectively using weak supervision signals through aggregated predictions. 
\end{enumerate}


\section{Related Work}
We identify three relevant lines of research: (1) zero-shot multi-label text classification; (2) weakly supervised single-label text classification; and (3) weakly supervised hierarchical multi-label text classification.

\paragraph{Zero-Shot Multi-Label Text Classification.} 
Zero-shot learning refers to a model's ability to recognize new labels that are not seen during training. This is typically achieved by learning semantic relationships between labels and input text through external linguistic knowledge bases \citep{0SHOT-TC}. Many zero-shot learning methodologies have been developed for single-label text classification \citep{0SHOT-TC, 3CosMul-TC, LTA, TE-Wiki}. 
The zero-shot multi-label text classification problem remains much less explored. 
Most existing work specializes in biomedical text classification, namely Automatic ICD (i.e., International Classification of Diseases) coding \citep{ZAGCNN, WGANZ}. \cite{ZAGCNN} use label descriptions to generate a feature vector for each label and employ a two layer graph convolutional network (GCN)~\citep{GCN} to encode the hierarchical label structure. \cite{WGANZ} propose a framework that exploits the hierarchical structure and label descriptions to construct relevant keyword sets through an adversarial generative model.  Although the setting is similar to ours in terms of problem definition, these methods rely heavily on substantial annotated data being available during training.


\paragraph{Weakly Supervised Single-Label Text Classification.} This problem assumes that there is no access to annotated data, but the full label set, with label names, and descriptions or keywords, is available. Usually, methods employ an iterative approach, building a vocabulary of keywords for each label and evaluating the overlap between the label vocabularies and input text content. \cite{LOTClass} use a pre-trained language model, BERT~\citep{BERT}, to generate a list of alternative words for each label. 
By comparing the text, the list of candidate replacements, it is determined which words in the text are potentially class-indicative. The method is effective, but computationally very expensive, since it requires running the pre-trained language model on every word in the labelled corpus. In addition, adaptation to a multi-label scenario requires training a binary classifier for each label. \cite{ConWea} argue that forming the keyword vocabulary for labels independent from the context of the input text makes it impossible for the model to differentiate between different usages of the same word. By using BERT~\citep{BERT} to build context vectors, they propose a method that can associate different meanings with different labels. \cite{ClassKG} observe that treating keywords of labels independently ignores the information embedded in keyword correlations. By building a keyword graph, the method they propose can take into account correlations using a graph neural network classifier. Existing techniques for weakly supervised text classification do not consider the multi-label classification task, and are challenging to extend to this setting because they do not account for label dependencies. The reliance on keywords limits the scope of their applicability.

\paragraph{Weakly Supervised Hierarchical Multi-label Text Classification.} For the setting where labels are organized in a hierarchical structure, most recent methods train a multi-label classifier in a supervised fashion using GCN~based architectures to encode the hierarchical relations \citep{HMLTC_GCN, HMLTC_RNN, HMLTC_Global}. Few methods have addressed the weakly supervised setting; an exception is the method proposed by \cite{TaxoClass},  which requires only label surface names in addition to category-subcategory relations represented as a directed acyclic graph. The method involves calculating a similarity score between each document-label pair using a pre-trained NLI model, and then traversing the hierarchy tree based on the similarity scores. The approach performs very well, but is limited to the setting where there is a strict hierarchy among the labels.


\section{Problem Statement}
Given a set of labels $\mathcal{L}= \{l, \varphi_l\}_{l=1}^{L}$ where $\varphi_l \in \Phi$ represents the textual description of the $l^{\text{th}}$ label and $L$ is the total number of unique labels, in multi-label text classification a sample $i$ is associated with input text content $\vartheta_i \in \Theta$ and a subset of labels $S_i \subset \mathcal{L}$. The aim is to design a classifier that can predict output labels by input text content $f: \Theta \mapsto \mathbb{S} = \mathcal{P}(\mathcal{L}) \backslash \emptyset$ where $\mathcal{P}(.)$ denotes the power set function. Let $\hat{S}_i$ denote the label subset predicted by the classifier for sample $i$ i.e., $f(\vartheta_i) = \hat{S}_i$. The quality of estimation can be evaluated by a variety of performance metrics that measure the similarity between the predicted $\hat{S}_i$ and the ground-truth label subset $S_i$. In our experiments we employ Hamming distance between binary label vectors as the primary performance metric, but we compare algorithms using multiple other assessment criteria. 

In this work, we consider three scenarios with different levels of supervision used for learning $f(\cdot)$ during training. Overall, the available supervision resources under consideration are defined as follows:
\begin{itemize}[leftmargin=*]
    \item[-] \textit{Contextual resources}
    \begin{enumerate}[leftmargin=*]
        \item \textit{Training data:} We are given a collection of samples $\{\vartheta_i\}_{i\in\mathcal{D}}$ without the ground-truth labels for learning $f(\cdot)$.
        \item \textit{Label descriptions:} Labels are meaningful, i.e., they are not a set of codes or indexes, and there is a sequence of words associated with each label that provides a description. We denote the set of possible labels and their corresponding descriptions by $\mathcal{L}= \{l, \varphi_l\}_{l=1}^{L}$, where $\varphi_l$ represents the textual description of the $l^{\text{th}}$ label and $L$ is the total number of unique labels. We assume that $L$ is known and covers both training and test samples.
        \item \textit{Average subset cardinality:} The expected number of labels per sample $\kappa = \mathbb{E}\left(|S|\right)$ is provided. 
        \item \textit{Label observation probabilities:} For each label, \emph{a priori} probability of inclusion of that label in a subset $\lambda_l = p(l \in S)$ is provided. 
        \item \textit{Annotated data:} There is a set of training data $\{\vartheta_i, S_i\}_{i\in\mathcal{D}_{\text{A}}}$ such that $\mathcal{D}_{\text{A}} \subset \mathcal{D}_{\text{train}}$ and $|\mathcal{D}_{\text{A}}| \ll |\mathcal{D}_{\text{train}}|$ with provided ground truth label subsets. 
    \end{enumerate}
    \item[-] \textit{External resources}
    \begin{enumerate}[leftmargin=*]
        \item \textit{Tokenizer:} We are given access to a pre-trained tokenization function with vocabulary $\mathcal{V}$ which is able to convert the input text content and label descriptions into a sequence of tokens, i.e., given an input text where $\tau \in \Theta \cup \Phi$, $f_{\text{tokenizer}}(\tau) = (t_1, \dots, t_s)$ such that $t_i \in \mathcal{V}$ and $s$ is the length of the input sequence.
        \item \textit{Language model:} We are given access to a pre-trained natural language inference model $f_{\text{NLI}}(\mathcal{H}, \mathcal{P})$ with vocabulary $\mathcal{V}$ that calculates true (entailment) $q$, undetermined (neutral) $\Tilde{q}$ or false (contradiction) $\Bar{q}$ probabilities of a hypothesis sequence $\mathcal{H} = (h_1, \dots, h_{s_h})$  where $h_i \in \mathcal{V}$, given a premise sequence $\mathcal{P} = (p_1, \dots, p_{s_p})$ where $p_j \in \mathcal{V}$ such that $q + \Tilde{q} + \Bar{q} = 1$.
        \item \textit{Word embeddings:} We are given access to a set of pre-trained $d$-dimensional word embeddings for the (tokens composing) label descriptions, i.e., $f_{\text{WE}}(t) = \mathbf{e}$ where $\mathbf{e} \in \mathbb{R}^d$ denotes embeddings of token $t \in \mathcal{V}$.  
    \end{enumerate}
\end{itemize}
We consider three different scenarios of supervision. In all scenarios, we are given \textit{training data}, \textit{label descriptions} and \textit{external resources}. The inputs of each scenario are summarized in Table \ref{tab:problem_settings}. We use a test set $\{j, \vartheta_j, S_j\}_{j\in \mathcal{D}_\text{{Test}}}$ such that $\mathcal{D} \cap \mathcal{D}_\text{{Test}} = \emptyset$ to evaluate the performance in all cases.
\begin{itemize}[leftmargin=*]
    \item[-] \textit{Annotation-Free:} In this scenario, we do not use any annotated data to learn the classifier but require supervision on average subset cardinality and label observation probabilities. 
    \item[-] \textit{Scarce-Annotation:} In this scenario, we have access to a small set of annotated data used for training, however average subset cardinality and label observation probabilities are not provided. 
    \item[-] \textit{Domain-Supervisor:} In this scenario, both a small set of annotated data and information regarding average subset cardinality and label observation probabilities are available. 
\end{itemize}
 
\begin{table}[t]
    \caption{Summary of problem settings with varying supervision signals}
    \label{tab:problem_settings}
    \centering
    \begin{tabular}{lcccccccc} \toprule
        Problem Setting & \multicolumn{5}{c}{Contextual Supervision} & \multicolumn{3}{c}{External Supervision } \\ 
        & $\mathcal{D}$ & $\mathcal{L}$ & $\kappa$ & $\lambda_l$ & $\mathcal{D}_{\text{A}}$ & $f_{\text{Tokenizer}}$ & $f_{\text{NLI}}$ & $f_{\text{WE}}$ \\ \midrule
        Annotation-Free     & \checkmark & \checkmark & \checkmark & \checkmark &   & \checkmark & \checkmark & \checkmark \\
        Scarce-Annotation   & \checkmark & \checkmark &   &   & \checkmark & \checkmark & \checkmark & \checkmark \\
        Domain-Supervisor  & \checkmark & \checkmark & \checkmark & \checkmark & \checkmark & \checkmark & \checkmark & \checkmark  \\ \bottomrule
    \end{tabular}
\end{table}

\section{Methodology}
Our proposed framework Balanced Neighbourhoods and Collective Loss (BNCL) for multi-label text classification consists of three components: (1) \textit{input transformation}, which maps input text into preliminary label predictions by natural language inference; (2) \textit{parameter preparation}, which involves calculation of a label dependency graph and mean data statistics; and (3) \textit{model update}, which updates the predictions obtained at the first stage. In this section, we share the details for each of these procedures. 

\subsection{Input Transformation}

The aim in natural language inference (NLI) is to determine whether a \textit{hypothesis} is true (entailment), undetermined (neutral) or false (contradiction) based on a given \textit{premise}. \cite{0SHOT-TC} formulate text classification as an NLI problem by treating input text as a premise and converting labels into hypotheses. To exemplify, let us consider a topic detection task on customer reviews with two possible topics of `product' and `delivery', for which the premise-hypothesis pairs could be developed as follows:
\begin{center}
    \begin{tabular}{lll}
        \underline{Premise:} & \underline{Hypothesis:} & \underline{Anticipated NLI output:} \\ 
        The material is very soft. & This review is about \textbf{product}. & entailment \\
        The material is very soft.  & This review is about \textbf{delivery}. & contradiction\\
        The parcel did not arrive on time. & This review is about \textbf{delivery}. & entailment
    \end{tabular}
\end{center}
When multiple true classes are not allowed (i.e., single-label classification), the entailment probabilities are compared and the largest is selected as the predicted class. In the multi-label scenario the entailment and contradiction probabilities are compared for each label independently in terms of entailment and contradiction probabilities, i.e., the problem is converted to binary relevance by ignoring neutral probabilities. For text classification, a predicted neutral for a label-specific hypothesis can be interpreted as the hesitancy of the language model to make a decision. The initial component of our proposed framework involves transforming the input text samples into a set of label-specific hypothesis probabilities using the NLI approach. This operation translates the input feature space into a 3-channel label space (i.e., entailment, neutral and contradiction probabilities). The procedure can be summarized as follows:
\begin{enumerate}[leftmargin=*]
    \item[] \textbf{Step 1: Convert corpus into premises, labels into hypotheses.} Following \cite{0SHOT-TC}, we build the hypothesis corresponding to label $l$ as \text{``This is about \{$\varphi_l$\}''}, where $\varphi_l$ is the label description, and we calculate the corresponding sequence of tokens $\mathcal{H}_l = f_{\text{Tokenizer}}(\text{``This is about \{$\varphi_l$\}''})$. Similarly, we treat each input text with content $\vartheta$ as a premise and calculate the corresponding sequence of tokens, $\mathcal{P} = f_{\text{Tokenizer}}(\vartheta)$.
    \item[] \textbf{Step 2: Query premise and hypothesis pairs.} Given a premise $\mathcal{P}$, we query $f_{\text{NLI}}(.)$ with all hypotheses $\{\mathcal{H}_l\}_{l \in \mathcal{L}}$ to calculate $\{(q_{l}, \tilde{q}_{l}, \bar{q}_{l})\}_{l \in \mathcal{L}}$. So now an input text is represented as a set of entailment, neutral and contradiction probabilities over labels:  
    \begin{equation}
        \vartheta \xmapsto[\mathcal{L} = \left\{l, \varphi_l\right\}_{l=1}^{L}]{f_{\text{NLI}}(.)} 
        \begin{array}{lll}
            \mathbf{q} = (q_1, \dots, q_L) \in [0, 1]^{L} \\
            \tilde{\mathbf{q}} = (\tilde{q}_1, \dots, \tilde{q}_L) \in [0, 1]^{L} \\
            \bar{\mathbf{q}} = (\bar{q}_1, \dots, \bar{q}_L) \in [0, 1]^{L}
        \end{array},
    \end{equation}
    where $q_{l}$, $\tilde{q}_{l}$ and $\bar{q}_{l}$ correspond to the probability of the hypothesis corresponding to label $l$ being true, undetermined, or false, respectively, given the premise $\vartheta$. Note that $q_{l} + \tilde{q}_{l} + \bar{q}_{l} = 1$. So if the representation is reduced to entailment and contradiction probabilities there is no loss of information. 
\end{enumerate}
Predicting entailment, neutral and contradiction probabilities for label-hypotheses by this procedure does not require any training with labelled data, and therefore does not incur any substantial annotation cost. However, the predictions rely only on external resources of language modelling and not tailored to the context associated with the dataset. We argue that the classification decisions can be enhanced by (1) modelling label dependencies with the help of label-specific features; and (2) incorporating supervision cheaper to obtain compared to mass amount of annotated data. 

\subsection{Parameter Preparation} 
Given label-hypothesis representations of inputs $\{(\mathbf{q}_{i}, \tilde{\mathbf{q}}_{i}, \bar{\mathbf{q}}_{i})\}_{i \in \mathcal{D}}$ as features, we learn an update function which requires (1) a signed label dependency graph $\mathcal{G} = (\mathcal{V}, \mathcal{E}^+, \mathcal{E}^-)$, where labels are represented as vertices $\mathcal{V} = \{1, \dots, L\}$, and edges are defined as tuples $(u, v) \in \mathcal{E}^+ (\in \mathcal{E}^-)$ indicating a positive (negative) dependency edge between labels $u$ and $v$; and (2) average subset cardinality $\kappa$ and label observation probabilities $\lambda_l$ as dataset-specific hyper-parameters. 

We form the label dependency graph using the following procedure. Given label descriptions $\mathcal{L}$ and word embeddings $f_{\text{WE}}$, for each label $l$ in $\mathcal{L} = \{l, \varphi_l\}$, the label description is tokenized and the corresponding label embedding is calculated as the average of the word embeddings of the tokens that compose the labels, i.e.,  $\mathbf{e}_l = \nicefrac{\sum_{t \in \mathcal{T}_l}\mathbf{e}_t}{s_l}$ where $\mathcal{T}_l = f_{\text{tokenizer}}(\varphi_l)$ denotes the sequence of tokens and $s_l = |\mathcal{T}_l|$ is the length of sequence. Afterwards, we calculate the cosine similarity between the embeddings of all label pairs $u, v \in \mathcal{L}$ by $d_{u, v} = \frac{\mathbf{e}_{u} . \mathbf{e}_{v}}{\|\mathbf{e}_{u}\|\|\mathbf{e}_{v}\|}$. Finally, the distances between label pairs are binarized by comparison to positive and negative edge thresholds $\delta^+$ and $\delta^-$, $\mathcal{E}^+ = \{(u, v): d_{u, v} \geq \delta^+\}$ and $\mathcal{E}^- = \{(u, v): d_{u, v} \leq \delta^-\}$. 


Average subset cardinality and label observation probabilities are assumed to be provided in \textit{annotation-free} and \textit{domain-supervisor} settings. In \textit{scarce-annotation}, we estimate both average statistics using the annotated set of data, i.e., $\hat{\kappa} = \frac{\sum_{i \in \mathcal{D}_{\text{A}}} |S_i|}{|\mathcal{D}_{\text{A}}|}$ and $\hat{\lambda_l} = \frac{\sum_{i \in \mathcal{D}_{\text{A}}} \mathbf{1}_{l \in S_i}}{|\mathcal{D}_{\text{A}}|}$. 


\subsection{Model Update}
Given the signed label dependency graph $\mathcal{G} = (\mathcal{V}, \mathcal{E}^+, \mathcal{E}^-)$, let $\mathbf{A}^{+} \in \{0, 1\}^{L \times L}$ and $\mathbf{A}^{-} \in \{0, 1\}^{L \times L}$ denote the adjacency matrices corresponding to positive $\mathcal{E}^+$ and negative $\mathcal{E}^-$ edges, respectively. 
\begin{equation}
    \mathbf{A}_{ij}^{\text{+}} = 
    \begin{cases}
        1, & \text{if there is a positive edge between $i$ and $j$} \\
        0,  & \text{otherwise}
    \end{cases}, 
    \mathbf{A}_{ij}^{-} = 
    \begin{cases}
        1, & \text{if there is a negative edge between $i$ and $j$} \\
        0,  & \text{otherwise}
    \end{cases}
\end{equation}
Since the label dependency graph is signed, finding the $k$-hop neighbourhoods for $k{>}1$ requires considering the interaction of negative and positive edges. \cite{SGCN} extend the graph convolutional network (GCN)~\citep{GCN} to signed networks based on balance theory, which states that a triad is balanced if and only if the number of negative edges is even; \textit{the friend of my friend is my friend and the enemy of my enemy is my friend.} Based on balance theory, we define the $k$-hop dependencies $\mathbf{D}^{(k,+)}$ and $\mathbf{D}^{(k,-)}$ recursively as follows:
\begin{align}
    \mathbf{D}^{(k,+)} &= \left(\mathbf{A}^{+}\right)^{\operatorname{T}} \mathbf{D}^{(k-1, +)} + \left(\mathbf{A}^{-}\right)^{\operatorname{T}} \mathbf{D}^{(k-1, -)} \\
    \mathbf{D}^{(k,-)} &= \left(\mathbf{A}^{+}\right)^{\operatorname{T}} \mathbf{D}^{(k-1, -)} + \left(\mathbf{A}^{-}\right)^{\operatorname{T}} \mathbf{D}^{(k-1, +)}
\end{align}
where $\mathbf{D}^{(1, +)} = \mathbf{A}^{+}$ and $\mathbf{D}^{(1, -)} = \mathbf{A}^{-}$. For $k=2$, this procedure corresponds to the following neighbourhoods:
\begin{align*}
    \mathbf{D}^{(1, +)} = & \mathbf{A}^+ & \rightarrow \text{friends} \\
    \mathbf{D}^{(1, -)} = & \mathbf{A}^- & \rightarrow \text{enemies} \\
    \mathbf{D}^{(2, +)} = & \left(\mathbf{A}^+\right)^{\operatorname{T}} \mathbf{A}^+ + \left(\mathbf{A}^-\right)^{\operatorname{T}} \mathbf{A}^- & \rightarrow \text{friends of friends + enemies of enemies} \\
    \mathbf{D}^{(2, -)} = & \left(\mathbf{A}^+\right)^{\operatorname{T}} \mathbf{A}^- + \left(\mathbf{A}^-\right)^{\operatorname{T}} \mathbf{A}^-  & \rightarrow \text{friends of enemies + enemies of friends}
\end{align*}

Finally, balanced neighbourhoods for label $v \in \mathcal{V}$ at hop $k \in \{1, \dots, K\}$ are formed as follows:
\begin{gather}
    \mathcal{N}_v^{(k,+)} = \{u: \mathbf{D}^{(k,+)}_{uv} > 0 \text{ for } u \in \mathcal{V} \}, \\
    \mathcal{N}_v^{(k,-)} = \{u: \mathbf{D}^{(k,-)}_{uv} > 0 \text{ for } u \in \mathcal{V} \}.
\end{gather}

\begin{figure}[t]
    \centering
    \includegraphics[width=0.9\textwidth]{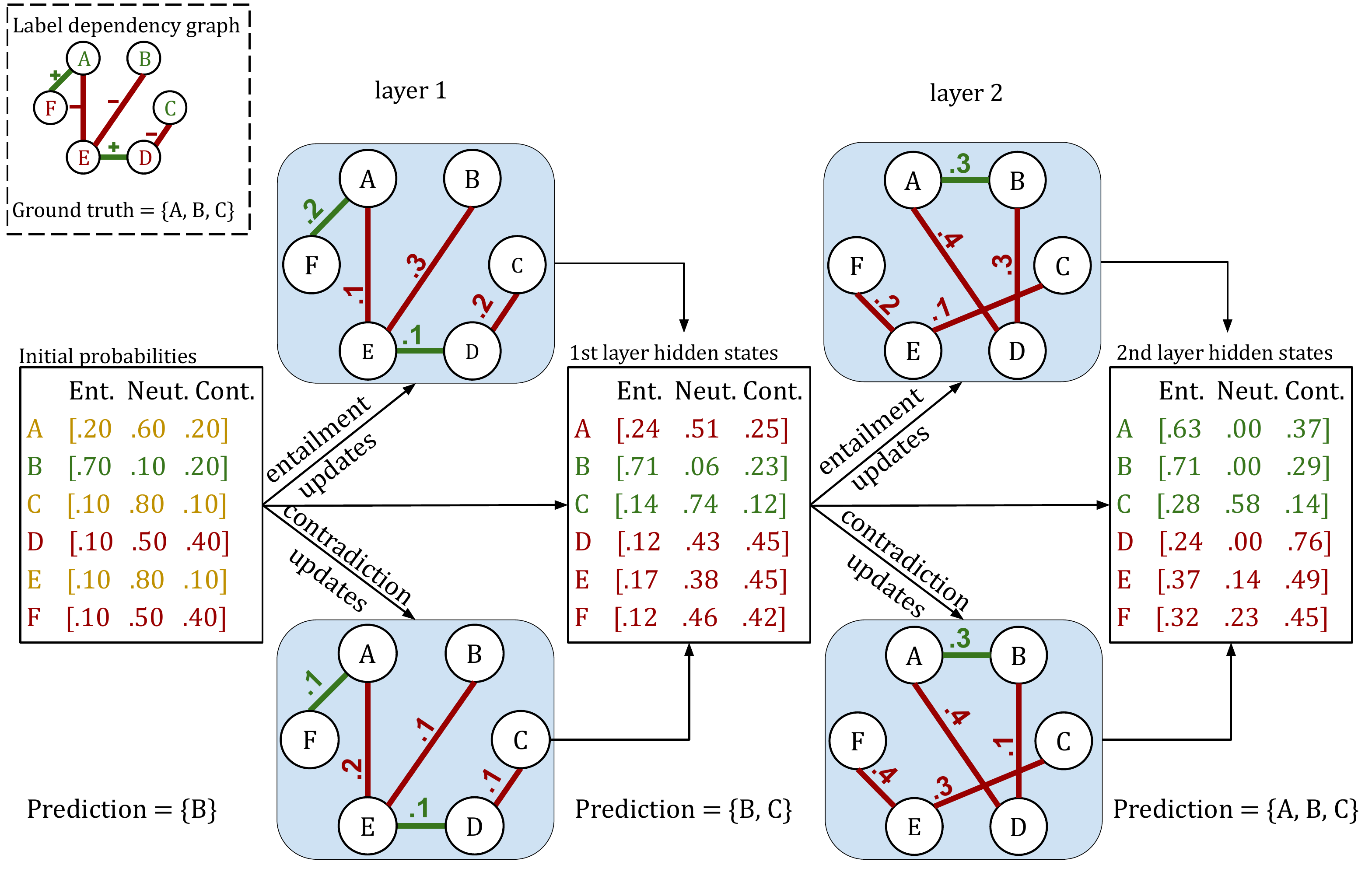}
    \caption{\small A toy example illustrating balanced neighbourhood update layers. Red and green edges represent negative and positive neighbourhoods at each layer. The initial and updated entailment, neutral, and contradiction probabilities are provided in tables for a sample with ground truth labels ``A'', ``B'' and ``C''. Edge attributes represent the learned weights that correspond to entailment (upper sequence) and contradiction (lower sequence) state updates. The initial predictions accepts ``B'', rejects ``D'' and ``F'' and does not make a decision on ``A'', ``C'' and ``E''. The first hob update with signed label dependency graph improves the prediction by adding ``C'' to predictions however rejects relevant label ``A''. At second hob, ``A'' is connected to one high entailment label ``B'' positively (as it is an enemy of enemy) and one high contradiction label ``D'' negatively (as it is friend of an enemy), which helps to improve the prediction by adding ``A'' to predictions. Note that the neutral probabilities drop through the updates. }
    \label{fig:toy_example}
\end{figure}

For each sample associated with entailment $\mathbf{q} = (q_1, \dots, q_L)$ and contradiction $\bar{\mathbf{q}} = (\bar{q}_1, \dots, \bar{q}_L)$ probabilities, we initialize its hidden representation by $\mathbf{h}^{(0)} = \mathbf{q}$ and $\bar{\mathbf{h}}^{(0)} = \bar{\mathbf{q}}$. Given the balanced neighbourhoods $\mathcal{N}_v^{(k,+)}$ and $\mathcal{N}_v^{(k,-)}$, the hidden states $\mathbf{h}^{(k)} = (h_1^{(k)}, \dots, h_L^{(k)})$ and $\bar{\mathbf{h}}^{(k)} = (\bar{h}_1^{(k)}, \dots, \bar{h}_L^{(k)})$ are updated at layer $k$ as follows:
\begin{align}
    \label{eq:layer_updates}
    h_v^{(k)} &= h_v^{(k-1)} + f_{\operatorname{ReLU}}\left(\sum_{v \in \mathcal{N}_v^{(k,+)}}\mathbf{W}_{uv}^{(k,+)} h_u^{(k-1)}\right) + f_{\operatorname{ReLU}}\left(\sum_{v \in \mathcal{N}_v^{(k,-)}}\overline{\mathbf{W}}_{uv}^{(k,-)} \bar{h}_u^{(k-1)}\right), \\    
    \bar{h}_v^{(k)} &= \bar{h}_v^{(k-1)} + f_{\operatorname{ReLU}}\left(\sum_{v \in \mathcal{N}_v^{(k,-)}}\mathbf{W}_{uv}^{(k,-)} h_u^{(k-1)}\right) + f_{\operatorname{ReLU}}\left(\sum_{v \in \mathcal{N}_v^{(k,+)}}\overline{\mathbf{W}}_{uv}^{(k,+)} \bar{h}_u^{(k-1)}\right),
\end{align}
where $\mathbf{W}^{(k,+)}, \mathbf{W}^{(k,-)}, \overline{\mathbf{W}}^{(k,+)}, \overline{\mathbf{W}}^{(k,-)} \in \mathbb{R}^{L \times L}$ are learnable weights and $f_{\operatorname{ReLU}}(.)$ denotes the Rectified Linear Unit function, i.e., $f_{\operatorname{ReLU}}(x) = \max(0, x)$. Figure~\ref{fig:toy_example} depicts the layer udpates.

During the testing phase, for a sample $i$, the set of predicted labels is determined by comparing the entailment and contradiction probabilities of each label independently, i.e., $\hat{S} = \{l: p_{i,l} > \bar{p}_{i,l}, \text{ for all } l \in \{1, \dots, L\}\}$. 

\paragraph{Loss Function.} Given the average subset cardinality $\kappa$ and label observation probabilities $\lambda_l$, the updates of entailment and contradiction probabilities are guided by a loss function composed of four components. Denote the final hidden states for a sample $i \in \mathcal{D}$ by $\mathbf{p}_i = \mathbf{h}_i^{(K)}$ and $\bar{\mathbf{p}}_i = \bar{\mathbf{h}}_i^{(K)}$, where $K$ is the total number of layers. The four components of the loss function are constructed as follows:
\begin{itemize}[leftmargin=*]
    \item[-] By definition entailment and contradiction are mutually exclusive events for each sample and each label. Therefore, the sum of the two cannot be greater than one, and as the sum becomes closer to zero, the neutral probability (hesitation to make a decision on the presence or absence of the label for a given sample) increases. Since hesitancy is undesirable, we  penalize the deviation of their summation from 1:
    \begin{equation}
        \mathbb{L}_{1} = \sum_{i \in \mathcal{D}} || \mathbf{p}_i + \bar{\mathbf{p}}_i - \mathbf{1} ||_2 .
    \end{equation}
    The entailment and contradiction representations are initialized to be non-negative. Since all the messages passed are non-negative, the hidden states remain non-negative (Equation \ref{eq:layer_updates}). This, together with the $\mathbb{L}_{1}$ term, motivates the model to protect the probability interpretation of hidden states while reducing the neutral probability.
    \item[-] For a specific training set, the classification decisions on training samples can impact each other. For example, a rare label may have very low entailment probability for all samples. Assuming the training data are representative, the samples to tag with that label can be selected by taking into account the expected observation probability. To ensure this, we penalize the difference between observed and expected probability for each label over training instances:
    \begin{equation}
        \mathbb{L}_{2} = \sum_{l = 1 }^{L}\left(|\mathcal{D}| \times \lambda_l - \sum_{i \in \mathcal{D}} \mathbf{1}_{p_{i,l} > \bar{p}_{i,l}}\right)^2 ,
    \end{equation}
    where $\mathbf{1}_{p_{i,l} > \bar{p}_{i,l}}$  is $1$ if entailment probability of label $l$ on sample $i$ is larger than its contradiction probability. 
    \item[-] It would be undesirable for some samples to have very high subset cardinality while others have zero. Therefore, we penalize the deviation from average subset cardinality for each sample:
    \begin{equation}
         \mathbb{L}_{3} = \sum_{i \in \mathcal{D}} \left(\kappa - \sum_{l = 1 }^{L} \mathbf{1}_{p_{i,l} > \bar{p}_{i,l}}\right)^2 .
    \end{equation}
    \item[-] In \textit{scarce-annotation} and \textit{domain-supervisor} settings, we have a small set of annotated data $\mathcal{D}_{\text{A}}$. Let $\mathbf{y}_i = (y_{i,1}, \dots, y_{i,L})$ denote the binary vector that corresponds to ground truth labels $S_i$ of sample $i$:
    \begin{equation}
         \mathbb{L}_{4} = \sum_{i \in \mathcal{D}_{\text{A}}} \sum_{i=1}^{L} - y_{i,l} \log(p_{i,l}) + (1-y_{i,l}) \log(\bar{p}_{i,l}) .
    \end{equation}
\end{itemize}
In order to make the term $\mathbf{1}_{p_{i,l} > \bar{p}_{i,l}}$ differentiable, we use a sharpened version of the sigmoid with a constant $C > 1$:
\begin{equation}
    \frac{1}{1 + e^{C \times (p_{i,l}-\bar{p}_{i,l})}} \approx \mathbf{1}_{p_{i,l} > \bar{p}_{i,l}}.
\end{equation}
The final loss function follows:
\begin{equation}
    \mathbb{L} = \mathbb{L}_{1} + \alpha_2 \mathbb{L}_{2} + \alpha_3 \mathbb{L}_{3} + \alpha_4  \mathbb{L}_{4},
\end{equation}where $\{\alpha_j\}_{j = 2}^{4}$ are hyperparameters used to scale the individual components of the loss function. 
\section{Experiments}
\paragraph{Datasets.} For our experiments we use two multi-label text classification datasets: Reuters21578\footnote{available at \url{https://archive.ics.uci.edu/ml/datasets/reuters-21578+text+categorization+collection}}~\citep{reuters}, which is a collection of newswire stories; and StackEx-Philosophy\footnote{available at \url{https://archive.org/download/stackexchange/philosophy.stackexchange.com.7z}}~\citep{stackex}, which is a collection of posts in Stack Exchange Philosophy forums. The dataset statistics are provided in Appendix \ref{appendix:dataset_statistics}. For both datasets, the label set is formed by the topics of the sample texts. For example, in Reuters21578 ``interest rates'' and ``unemployment'' are label descriptions, and in StackEx-Philosophy ``ethics'' and ``skepticism'' are label descriptions. 

\paragraph{Metrics.} In addition to Hamming accuracy (HA), we use example based F1 score (ebF1), subset accuracy (ACC), micro-averaged F1 score (miF1), and macro-averaged F1 score (maF1) as metrics to evaluate the performance of our method. Subset accuracy measures the fraction of times that an algorithm identifies the correct subset of labels for each instance. The example-based F1 score is aggregated over samples and the macro-averaged F1 score over labels. The micro-averaged F1 score takes the average of the F1 score weighted by the contribution of each label, and thus takes label imbalance into account. Expressions for the metrics are provided in equations \ref{eq:ACC} - \ref{eq:maF1} in Appendix \ref{appendix:metrics}. 

\paragraph{Baselines.} To the best of our knowledge, the problem settings under consideration in this study have not been explored directly in the literature. The weakly supervised text classification problem (in which only label descriptions are given) has been studied primarily in the single-label classification context \citep{LOTClass, ConWea, ClassKG, WDDC}. Some works impose constraints such as the requirement that all label indicative keywords should be seen in the corpus. We attempted to adapt one of the state-of-the-art methods, LOTClass~\citep{LOTClass}, to the multi-label scenario, but encountered errors regarding these constraints. Execution was only possible if more than half of the labels in the dataset were excluded. TaxoClass~\citep{TaxoClass} considers multi-label classification with no annotated data, but it requires a hierarchy tree which represents category-subcategory types of relations between labels. Furthermore, all labels must be aligned exactly with the hierarchy tree. We compare to:
\begin{itemize}[leftmargin=*]
\item[-] \textbf{0Shot-TC~\citep{0SHOT-TC}} (multi-label version), which uses the NLI formulation of the text classification task. Estimated entailment/contradiction probabilities per label are used directly to make classification decisions. Since the same formulation is used for our input transformation, this comparison reveals the impact of our ``model update'' module on multi-label classification performance in contrast to using raw language model output. 
\item[-] \textbf{ML-KNN~\citep{MLKNN}}, a multi-label classifier originally designed for the supervised setting, which finds the nearest examples to a test class using the k-Nearest Neighbors algorithm and then selects the assigned labels using Bayesian inference.
\item[-] \textbf{ML-ARAM~\citep{MLARAM}}, a multi-label classifier designed for the supervised setting, which use Adaptive Resonance Theory (ART) based clustering and Bayesian inference to calculate label probabilities. 
\end{itemize}

\paragraph{Experimental settings.} We examine performance in three different experimental settings, as identified in the problem statement. In ``Annotation-Free'', no annotated data is available for training, but we assume knowledge of average subset cardinality and label observation probabilities. In ``Scarce Annotation'', a small annotated dataset is available. In our experiments, the annotated dataset size is set to $L$. In ``Domain-Supervisor'', in addition to the annotated dataset, knowledge concerning the average subset cardinality and label observation probabilities is available. 

\begin{table}[t]
\caption{Comparison between the proposed method and 0Shot-MLTC in the Annotation-Free setting. The table also shows performance for the Scarce-Annotation and Domain-Supervisor settings. All results are calculated over 10 random initialization on the original train-test data splits.}
\label{tab:settings_comparison}
\begin{center}
\resizebox{\textwidth}{!}{
\begin{tabular}{@{}ccccccc|ccccc@{}}
\toprule
          & & \multicolumn{5}{c}{\textit{Reuters21578}}           & \multicolumn{5}{c}{\textit{StackEx-Philosophy}} \\ 
          &  & ACC    & HA     & ebF1   & miF1   & maF1   & ACC   & HA   & ebF1   & miF1   & maF1  \\ \midrule
0Shot-MLTC &  & 0.0834 & 0.9799 & 0.2778 & 0.2981 & 0.1844 & 0.001 & 0.8802 & 0.0924 & 0.0665 & 0.1528 \\ \midrule
\makecell{BNCL \\ \textit{Annotation-Free}} & \makecell{mean \\ \textit{std}} & \makecell{0.3159 \\ \textit{0.0446}} & \makecell{0.9917 \\ \textit{0.0007}} & \makecell{0.4613 \\ \textit{0.0393}} & \makecell{0.5053 \\ \textit{0.0343}} & \makecell{0.2184 \\ \textit{0.0034}} &  \makecell{0.0382 \\ \textit{0.0035}} & \makecell{0.9902	\\ \textit{0.0001}} & \makecell{0.2119 \\ \textit{0.0042}} & \makecell{0.2423 \\ \textit{0.0035}} & \makecell{0.2292 \\ \textit{0.0054}} \\ \midrule
\makecell{BNCL \\ \textit{Scarce-Annotation}} & \makecell{mean \\ \textit{std}} & \makecell{\textbf{0.5083} \\ \textit{0.0134}} & \makecell{\textbf{0.9944} \\ \textit{0.0001}} & \makecell{0.6318 \\ \textit{0.0131}} & \makecell{0.6595 \\ \textit{0.0101}} & \makecell{0.2340 \\ \textit{0.0091}} & \makecell{0.0547 \\ \textit{0.0030}} & \makecell{\textbf{0.9913} \\ \textit{0.0002}} & \makecell{\textbf{0.2655} \\ \textit{0.0055}} & \makecell{\textbf{0.3024} \\ \textit{0.0047}} & \makecell{0.2304 \\ \textit{0.0091}} \\ \midrule
\makecell{BNCL \\ \textit{Domain-Supervisor}} & \makecell{mean \\ \textit{std}} & \makecell{0.5078 \\ \textit{0.0120}} & \makecell{\textbf{0.9944} \\ \textit{0.0001}} & \makecell{\textbf{0.6320} \\ \textit{0.0145}} & \makecell{\textbf{0.6606} \\ \textit{0.0108}} & \makecell{\textbf{0.2353} \\ \textit{0.0085}} & \makecell{\textbf{0.0551} \\ \textit{0.0028}} & \makecell{\textbf{0.9913} \\ \textit{0.0001}} & \makecell{0.2648 \\ \textit{0.0058}} & \makecell{0.3021 \\ \textit{0.0059}} & \makecell{\textbf{0.2309} \\ \textit{0.0090}} \\ \bottomrule
\end{tabular}
}
\end{center}
\end{table}


\paragraph{Implementation.} We transform the input using the pre-trained model BART~\citep{BART} and its corresponding tokenizer, which is fine-tuned on a large corpus, MNLI~\citep{MNLI}, composed of hypothesis-premise pairs. For both datasets, the maximum sequence length of the tokenizer is set to 128. We use GloVe~\citep{GloVe} to generate word embeddings to calculate the label graph from the label descriptions. The positive and negative edge thresholds $\delta^{+}$ and $\delta^{-}$ that control label graph density are set by top-bottom percentiles of the overall distribution of the distances between label embeddings. For each dataset, it is selected from the following list of percentile pairs $[(5\%, 95\%), (10\%, 90\%), (30\%, 70\%)]$. When the average subset cardinality and label observation probabilities are assumed to be provided, they are calculated based on the whole set of training data. In the scarce-annotation and domain-supervisor settings, the size of the annotated dataset is $L$ (i.e., $|\mathcal{D}_{\text{A}}| = |\mathcal{L}| = L$.). The annotated examples are randomly selected from the training set. The sigmoid sharpening factor $C$ is set to 10. The procedure to select this value was: (1) sample a small set of examples; (2) compare their entailment and contradiction probabilities to determine predicted label subsets; (3) calculate the sharpened sigmoid function value, successively increasing the integer $C$ by one; and (4) choose the smallest integer $C$ such that the output for all sample/predicted label pairs is greater than 0.9999. The loss function scaling factors $\{\alpha_j\}_{j=2}^{4}$ are tuned using grid search over $\alpha_2, \alpha_3 \in \{0.1, 0.5, 1\}$ and $\alpha_4 \in \{1, 10, 100\}$. The selected values for both datasets are at $\alpha_2 = 0.1, \alpha_3 = 0.5, \alpha_4 = 100$. If not stated otherwise, the number of update layers is set to $2$ because a smaller number caused validation performance to be too sensitive to label graph density, and a greater number reduced the performance on the validation data. The model is trained with a batch size of 128 for 30 epochs as it is observed that validation performance does not improve after 30 epochs. The Adam~\citep{adamoptimizer} optimizer is used to compute gradients and update parameters with the initial learning rate of $1 \times 10^{-3}$ and beta coefficients of $(0.8, 0.9)$. The learning rate is updated with a step size 10 for a 10\% decay rate. The results for ML-KNN and ML-ARAM are obtained by implementations provided in the scikit-learn library~\citep{scikit-learn}. These algorithms are both trained using the full sets of training data. In the comparison with these supervised algorithms, we train our algorithm using $50\%$ of the annotations.

\begin{table}[t]
\caption{Comparison with supervised baseline methods. BNCL outperforms the baselines with in the 100\% Annotation setting and achieves equivalent performance for 50\% Annotation. Performance degradation as the annotation level decreases is graceful.}
\label{tab:supervised_comparison}
\begin{center}
\begin{tabular}{@{}lccccc|ccccc@{}}
\toprule
          & \multicolumn{5}{c}{\textit{Reuters21578}}           & \multicolumn{5}{c}{\textit{StackEx-Philosophy}} \\ 
          & ACC    & HA     & ebF1   & miF1   & maF1   & ACC   & HA   & ebF1   & miF1   & maF1  \\ \midrule
ML-KNN    & 0.6513 & 0.9956 & 0.7101 & 0.7228 & 0.2536 & \textbf{0.0904} & \textbf{0.9925} & 0.2866 & 0.3198 & 0.0993 \\
ML-ARAM   & 0.4742 & 0.9923 & 0.6734 & 0.6265 & 0.1633 & 0.0622     & 0.9888    & 0.2045      & 0.1796      & 0.0075     \\
BNCL-100\% & \textbf{0.6674} & \textbf{0.9961} & \textbf{0.7772} & \textbf{0.7720} & \textbf{0.2784} & 0.0803 & 0.9917 & \textbf{0.3394} & \textbf{0.3749} & 0.2387 \\
BNCL-50\% & 0.6039 & 0.9954 & 0.7120 & 0.7286 & 0.2589 & 0.0723 & 0.9917 & 0.3283 & 0.3642 & \textbf{0.2586} \\
BNCL-20\% & 0.5618 & 0.9949 & 0.6881 & 0.7028 & 0.2459 & 0.0643 & 0.9916 & 0.2948 & 0.3241 & 0.2418 \\
BNCL-5\% & 0.3784 & 0.9922 & 0.5763 & 0.5853 & 0.2439 & 0.0331 & 0.9880 & 0.2696 & 0.2813 & 0.2266 \\\bottomrule
\end{tabular}
\end{center}
\end{table}

\paragraph{Comparison with 0Shot-MLTC.} Table \ref{tab:settings_comparison} compares the performance of 0Shot-TC adapted to multi-label scenario and the performance of our proposed method, BNCL. We examine how BNCL performs in three settings. In the annotation-free setting, we see that BNCL achieves much better performance for all metrics. This indicates how valuable it is to construct the signed label dependency graph and use it to update the embeddings using the signed graph convolution network. Moving from the annotation-free to the scarce-annotation setting,  subset accuracy improves by 63\% and 43\%, example-based F1 score by 37\% and 25\% and micro-averaged F1 score by 31\% and 25\%, on Reuters21578 and StackEx-Philosophy, respectively. This shows that having a small set of annotated data is very helpful. There is no meaningful performance difference between the scarce-annotation and domain-supervisor settings, which suggests that when a small amount of annotation is available, there is no need for supervision in terms of average label subset cardinality and label observation probabilities. 

\paragraph{Comparison to Supervised Learning Methods.} Table \ref{tab:supervised_comparison} compares the results of two supervised baseline methods which are trained with full set of training data to BNCL, which is trained with various ratios of data annotation. We find that BNCL with only 50\% annotation achieves similar performance to the supervised baseline methods. We also find that BNCL with 100\% performs better than the supervised methods, even though it is designed for the limited annotation settings and uses less annotated data in this setting.  

\begin{figure}[t]
    \centering
    \textit{Reuters21578} \\ 
    \includegraphics[width=0.32\textwidth]{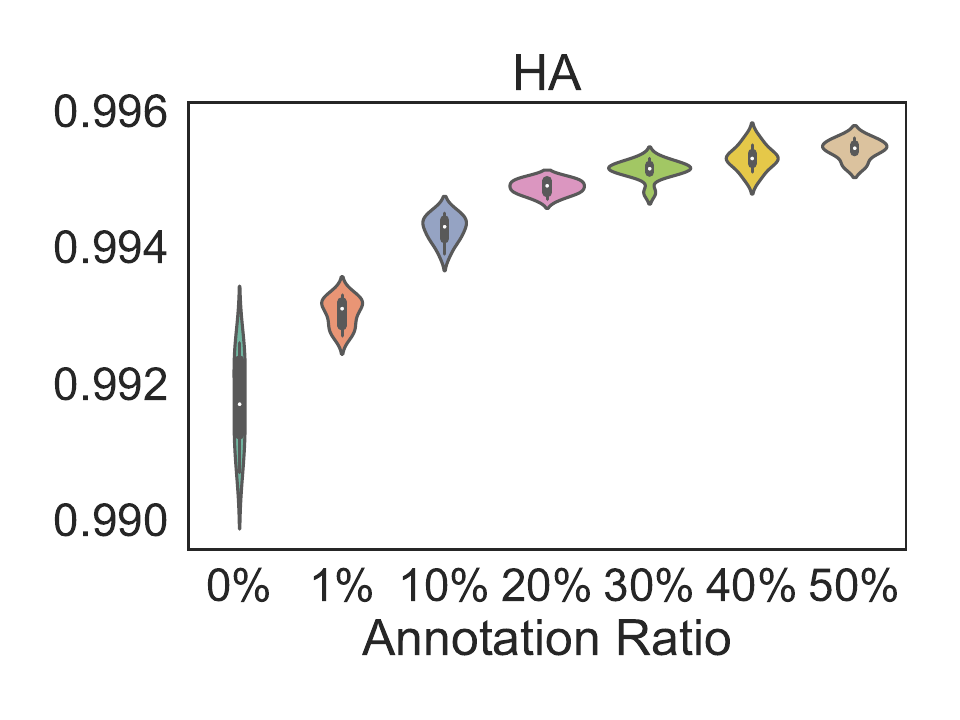}
    \includegraphics[width=0.32\textwidth]{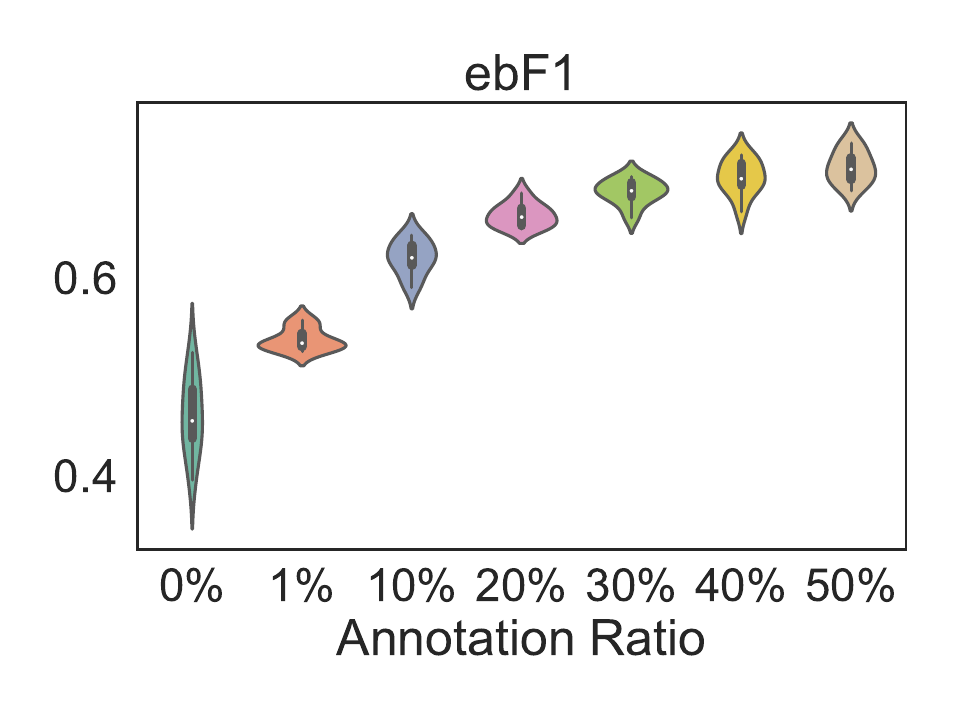} 
    \includegraphics[width=0.32\textwidth]{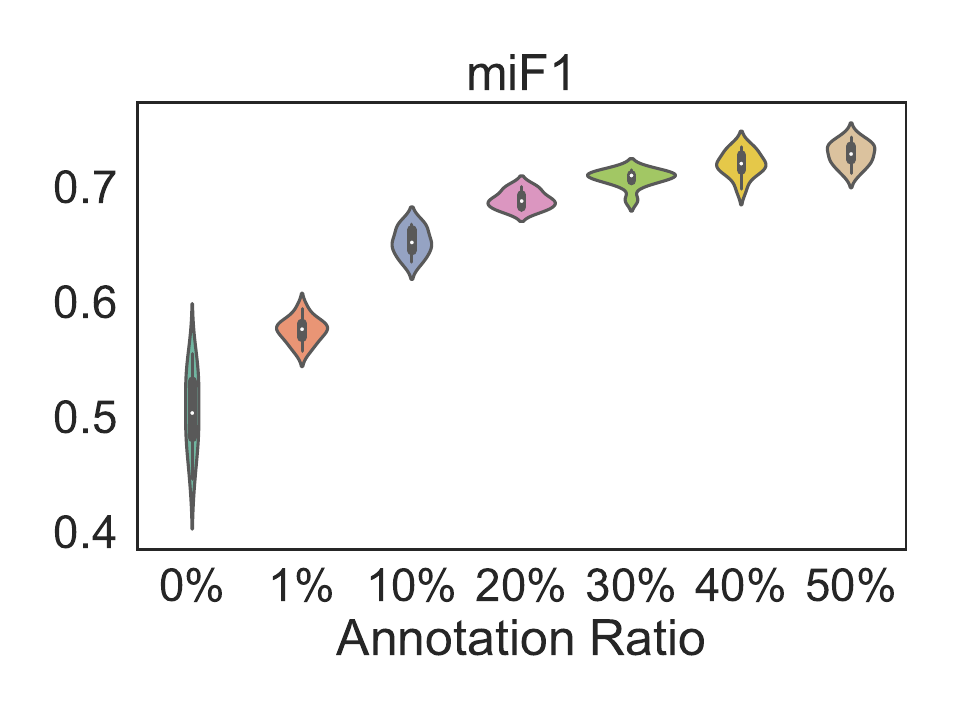} \\
    \textit{StackEx-Philosophy} \\ 
    \includegraphics[width=0.32\textwidth]{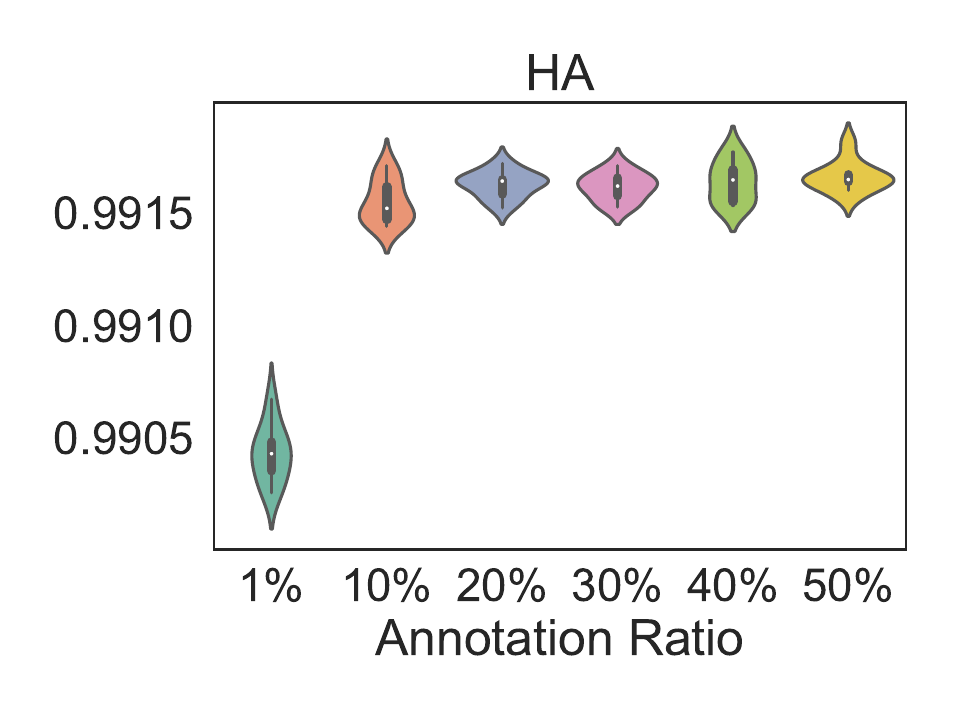}
    \includegraphics[width=0.32\textwidth]{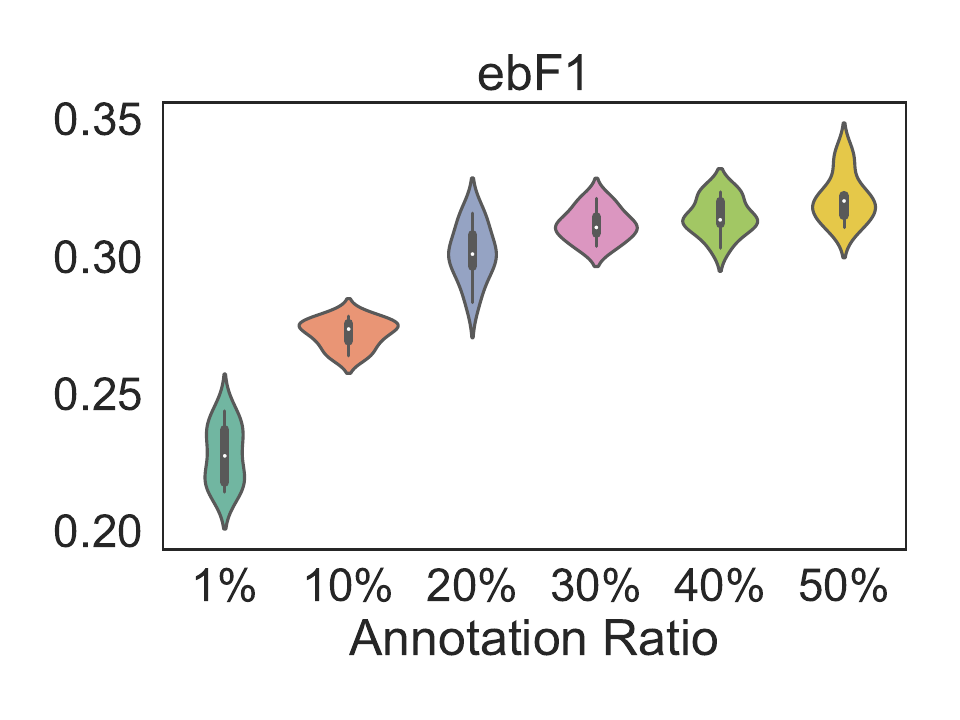}
    \includegraphics[width=0.32\textwidth]{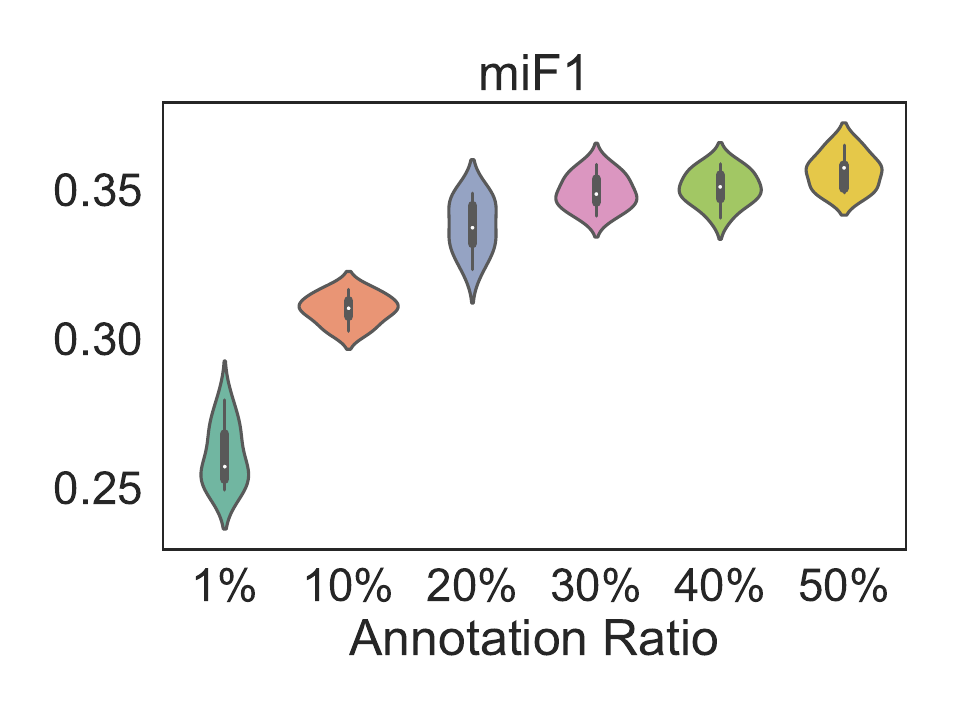} 
    \caption{Sensitivity study with different amounts of annotated data}
    \label{fig:sensitivity_annotation_level}
\end{figure}

\paragraph{Sensitivity Study - Annotation Level.} In order to observe the impact of the amount of annotated data, we perform a sensitivity study by changing the level of annotation in the domain-supervisor setting. In Figure \ref{fig:sensitivity_annotation_level}, the performance for different levels of annotation is presented for both datasets in terms of Hamming accuracy, example-based F1 score, and micro-averaged F1 score. The pattern observable for all three metrics on both datasets is that even a small set of annotated data (as little as 1\% of the training data) is capable of improving the performance. But as the amount of annotation increases, the improvement diminishes, and beyond 30\%, there is much less value in further annotation.

\paragraph{Ablation Study - Loss Function Components.} In order to understand the impact of individual loss components, we perform an ablation study. We examine the impact of removing $\mathbb{L}_2$, which targets matching of the label observation probabilities, and $\mathbb{L}_3$, which aims to balance the subset cardinality. The study is conducted in the annotation-free setting on the StackEx-Philosophy dataset. Table \ref{tab:ablation_stackex} compares the results of the original loss function to configurations (1) without the $\mathbb{L}_2$ component; (2) without the $\mathbb{L}_3$ component; and (3) without both. The results show that using label observation probabilities to guide the update of label-hypothesis probabilities significantly improves the performance. On the other hand, removing the $\mathbb{L}_3$ component that is associated with balancing subset cardinality results in a relatively small deterioration in accuracy, and actually improves the example based and micro-averaged F1 scores. This is likely due to the power law distribution that label subset cardinalities follow in the StackEx-Philosophy dataset. Constraining each sample's label subset cardinality to an expected level may harm the example based performance especially in terms of most frequent labels (note that miF1 weighs infrequent labels less compared to maF1). When both components are missing the performance drops dramatically. This indicates that either component can provide valuable regularizing information, but without both, the model update module is drawn towards poor representations. 

\begin{table}[]
\caption{Ablation study for loss function components on StackEx-Philosophy dataset}
\label{tab:ablation_stackex}
\small
\begin{center}
\begin{tabular}{@{}ccccccc@{}}
\toprule
                   &      & ACC    & HA     & ebF1   & miF1   & maF1   \\ \midrule
BNCL               & mean & \textbf{0.0382} & \textbf{0.9902} & 0.2119 & 0.2423 & \textbf{0.2292} \\
\textit{Original}           & \textit{std}  & \textit{0.0035} & \textit{0.0001} & \textit{0.0042} & \textit{0.0035} & \textit{0.0054} \\ \midrule
BNCL               & mean & 0.0236 & 0.9899 & 0.1921 & 0.2222 & 0.2267 \\
\textit{Removing $\mathbb{L}_2$}        & \textit{std}  & \textit{0.0028} & \textit{0.0001} & \textit{0.0048} & \textit{0.0040} & \textit{0.0063} \\ \midrule
BNCL               & mean & 0.0347 & 0.9865 & \textbf{0.2373} & \textbf{0.2458} & 0.2025 \\
\textit{Removing $\mathbb{L}_3$}        & \textit{std}  & \textit{0.0085} & \textit{0.0018} & \textit{0.0136} & \textit{0.0032} & \textit{0.0083} \\ \midrule
BNCL               & mean & 0.0004 & 0.7996 & 0.0509 & 0.0374 & 0.0736 \\
\textit{Removing $\mathbb{L}_2$ and $\mathbb{L}_3$} & \textit{std}  & \textit{0.0005} & \textit{0.0092} & \textit{0.0027} & \textit{0.0016} & \textit{0.0034} \\ \bottomrule
\end{tabular}
\end{center}
\end{table}
\section{Conclusion}
\paragraph{Summary and Contributions.} In this study, we propose a framework for multi-label text classification in the absence of strong supervision signals. Our framework performs transfer learning using external knowledge bases, and exploits the benefits of modelling the dependencies between labels in order to focus the external supervision on domain-specific properties of the data. We project input text onto a label-hypothesis probability space using a pre-trained language model and then update representations using the guidance of label dependencies and aggregated predictions over the training data. To the best of our knowledge this is the first work that considers weakly supervised multi-label text classification problem when the label space is not strictly structured according to a set of label hierarchies. 

\paragraph{Limitations and Future Work.} Extreme Multi-label Learning (XML) involves finding the most relevant subset of labels for each data point from an extremely large label set. The number of labels can scale to thousands or millions. Using our method in an extreme classification setting would be infeasible due to the computational overhead of the input transformation process (we need to calculate probabilities for every candidate label for every text example). One future work direction we would like to explore is developing an active learning based framework in order to select the labels to query for each input text. Another limitation associated with the proposed method is the inability to handle noise in the values provided by a domain-supervisor with respect to average subset cardinality or label observation probabilities. We also do not take into account any uncertainty in the signed label graph constructed from the label descriptions. Therefore, desirable follow up work involves improving our methodology by incorporating Bayesian approaches to account for uncertainty in the estimated parameters and label dependency graph.

\newpage
\bibliography{references}

\begin{thebibliography}{33}
\providecommand{\natexlab}[1]{#1}
\providecommand{\url}[1]{\texttt{#1}}
\expandafter\ifx\csname urlstyle\endcsname\relax
  \providecommand{\doi}[1]{doi: #1}\else
  \providecommand{\doi}{doi: \begingroup \urlstyle{rm}\Url}\fi

\bibitem[Benites \& Sapozhnikova(2015)Benites and Sapozhnikova]{MLARAM}
F.~Benites and E.~Sapozhnikova.
\newblock Haram: A hierarchical aram neural network for large-scale text
  classification.
\newblock In \emph{Proc. IEEE Int. Conf. Data Mining Workshop}, pp.\  847--854,
  2015.

\bibitem[Charte \& Charte(2015)Charte and Charte]{stackex}
Francisco Charte and David Charte.
\newblock Working with multilabel datasets in r: The mldr package.
\newblock \emph{R J.}, 7:\penalty0 149, 2015.

\bibitem[Derr et~al.(2018)Derr, Ma, and Tang]{SGCN}
Tyler Derr, Yao Ma, and Jiliang Tang.
\newblock Signed graph convolutional network.
\newblock In \emph{IEEE Int. Conf. Data Mining (ICDM)}, pp.\  1066–1075,
  2018.

\bibitem[Devlin et~al.(2019)Devlin, Chang, Lee, and Toutanova]{BERT}
Jacob Devlin, Ming-Wei Chang, Kenton Lee, and Kristina Toutanova.
\newblock {BERT}: Pre-training of deep bidirectional transformers for language
  understanding.
\newblock In \emph{Proc. Association for Computational Linguistics (ACL)}, pp.\
   4171--4186, 2019.

\bibitem[Ding et~al.(2022)Ding, Yang, Deng, Zhang, and Roth]{TE-Wiki}
Hantian Ding, Jinrui Yang, Yuqian Deng, Hongming Zhang, and Dan Roth.
\newblock Towards open-domain topic classification.
\newblock In \emph{Proc. NAACL - Human Language Technologies}, pp.\  90--98,
  2022.

\bibitem[Gururangan et~al.(2019)Gururangan, Dang, Card, and Smith]{SemiTC}
Suchin Gururangan, Tam Dang, Dallas Card, and Noah~A. Smith.
\newblock Variational pretraining for semi-supervised text classification.
\newblock In \emph{Proc. Association for Computational Linguistics (ACL)},
  2019.

\bibitem[Huang et~al.(2019)Huang, Chen, Liu, Chen, Huang, Liu, Zhao, Zhang, and
  Wang]{HMLTC_RNN}
Wei Huang, Enhong Chen, Qi~Liu, Yuying Chen, Zai Huang, Yang Liu, Zhou Zhao,
  Dan Zhang, and Shijin Wang.
\newblock Hierarchical multi-label text classification: An attention-based
  recurrent network approach.
\newblock In \emph{In. Proc. Conf. Information and Knowledge Management
  (CIKM)}, 2019.

\bibitem[Kingma \& Ba(2015)Kingma and Ba]{adamoptimizer}
D.~P. Kingma and J.~Ba.
\newblock Adam: A method for stochastic optimization.
\newblock In \emph{Proc. Int. Conf. Learning Representations (ICLR)}, 2015.

\bibitem[Kipf \& Welling(2017)Kipf and Welling]{GCN}
T.~N. Kipf and M.~Welling.
\newblock Semi-supervised classification with graph convolutional networks.
\newblock In \emph{Proc. Int. Conf. Learning Representations (ICLR)}, 2017.

\bibitem[Lewis et~al.(2004)Lewis, Yang, Rose, and Li]{reuters}
D.~D. Lewis, Y.~Yang, T.~G. Rose, and F.~Li.
\newblock Rcv1: A new benchmark collection for text categorization research.
\newblock \emph{J. Machine Learning Research}, 5:\penalty0 361–397, 2004.

\bibitem[Lewis et~al.(2020)Lewis, Liu, Goyal, Ghazvininejad, Mohamed, Levy,
  Stoyanov, and Zettlemoyer]{BART}
Mike Lewis, Yinhan Liu, Naman Goyal, Marjan Ghazvininejad, Abdelrahman Mohamed,
  Omer Levy, Veselin Stoyanov, and Luke Zettlemoyer.
\newblock {BART}: Denoising sequence-to-sequence pre-training for natural
  language generation, translation, and comprehension.
\newblock In \emph{Proc. Association for Computational Linguistics (ACL)}, pp.\
   7871--7880, 2020.

\bibitem[Liu et~al.(2017)Liu, Chang, Wu, and Yang]{XML-CNN}
Jingzhou Liu, Wei-Cheng Chang, Yuexin Wu, and Yiming Yang.
\newblock Deep learning for extreme multi-label text classification.
\newblock In \emph{Proc. Int. ACM SIGIR Conf. Research and Development in
  Information Retrieval}, pp.\  115–124, 2017.

\bibitem[Mekala \& Shang(2020)Mekala and Shang]{ConWea}
Dheeraj Mekala and Jingbo Shang.
\newblock Contextualized weak supervision for text classification.
\newblock In \emph{Proc. Association for Computational Linguistics (ACL)}, pp.\
   323--333, 2020.

\bibitem[Meng et~al.(2020)Meng, Zhang, Huang, Xiong, Ji, Zhang, and
  Han]{LOTClass}
Yu~Meng, Yunyi Zhang, Jiaxin Huang, Chenyan Xiong, Heng Ji, Chao Zhang, and
  Jiawei Han.
\newblock Text classification using label names only: A language model
  self-training approach.
\newblock In \emph{Proc. Conf. Empirical Methods Natural Language Processing
  (EMNLP)}, pp.\  9006--9017, 2020.

\bibitem[Nam et~al.(2017)Nam, Loza~Menc\'{\i}a, Kim, and F\"{u}rnkranz]{MLRNN}
J.~Nam, E.~Loza~Menc\'{\i}a, H.~J. Kim, and J.~F\"{u}rnkranz.
\newblock Maximizing subset accuracy with recurrent neural networks in
  multi-label classification.
\newblock In \emph{Proc. Advances in Neural Information Processing Systems
  (NeurIPS)}, pp.\  5413--5423, 2017.

\bibitem[Ozmen et~al.(2022)Ozmen, Zhang, Wang, and Coates]{MrMP}
M.~Ozmen, H.~Zhang, P.~Wang, and M.~Coates.
\newblock Multi-relation message passing for multi-label text classification.
\newblock In \emph{Proc. IEEE Int. Conf. Acoustics, Speech and Signal
  Processing (ICASSP)}, 2022.

\bibitem[Pedregosa et~al.(2011)Pedregosa, Varoquaux, Gramfort, Michel, Thirion,
  Grisel, Blondel, Prettenhofer, Weiss, Dubourg, Vanderplas, Passos,
  Cournapeau, Brucher, Perrot, and Duchesnay]{scikit-learn}
F.~Pedregosa, G.~Varoquaux, A.~Gramfort, V.~Michel, B.~Thirion, O.~Grisel,
  M.~Blondel, P.~Prettenhofer, R.~Weiss, V.~Dubourg, J.~Vanderplas, A.~Passos,
  D.~Cournapeau, M.~Brucher, M.~Perrot, and E.~Duchesnay.
\newblock Scikit-learn: machine learning in {p}ython.
\newblock \emph{J. Machine Learning Research}, 12:\penalty0 2825--2830, 2011.

\bibitem[Peng et~al.(2018)Peng, Li, He, Liu, Bao, Wang, Song, and
  Yang]{HMLTC_GCN}
Hao Peng, Jianxin Li, Y.~He, Yaopeng Liu, Mengjiao Bao, L.~Wang, Y.~Song, and
  Qiang Yang.
\newblock Large-scale hierarchical text classification with recursively
  regularized deep graph-cnn.
\newblock In \emph{Proc. World Wide Web Conf.}, 2018.

\bibitem[Pennington et~al.(2014)Pennington, Socher, and Manning]{GloVe}
Jeffrey Pennington, Richard Socher, and Christopher~D. Manning.
\newblock Glove: Global vectors for word representation.
\newblock In \emph{Proc. Conf. Empirical Methods Natural Language Processing
  (EMNLP)}, 2014.

\bibitem[Rios \& Kavuluru(2018)Rios and Kavuluru]{ZAGCNN}
Anthony Rios and Ramakanth Kavuluru.
\newblock Few-shot and zero-shot multi-label learning for structured label
  spaces.
\newblock In \emph{Proc. Conf. Empirical Methods Natural Language Processing
  (EMNLP)}, pp.\  3132--3142, 2018.

\bibitem[Shen et~al.(2021)Shen, Qiu, Meng, Shang, Ren, and Han]{TaxoClass}
Jiaming Shen, Wenda Qiu, Yu~Meng, Jingbo Shang, Xiang Ren, and Jiawei Han.
\newblock Taxoclass: Hierarchical multi-label text classification using only
  class names.
\newblock In \emph{Proc. NAACL - Human Language Technologies}, pp.\
  4239--4249, 2021.

\bibitem[Song et~al.(2020)Song, Zhang, Sadoughi, Xie, and Xing]{WGANZ}
Congzheng Song, Shanghang Zhang, Najmeh Sadoughi, Pengtao Xie, and Eric Xing.
\newblock Generalized zero-shot text classification for icd coding.
\newblock In \emph{Proc. Int. Joint Conference on Artificial Intelligence
  (IJCAI)}, pp.\  4018--4024, 2020.

\bibitem[Tsoumakas \& Katakis(2007)Tsoumakas and Katakis]{mlcReview}
G.~Tsoumakas and I.~Katakis.
\newblock Multi-label classification: an overview.
\newblock \emph{Int. J. Data Warehousing and Mining}, 3:\penalty0 1--13, 2007.

\bibitem[Varma(2018)]{xmlReview}
Manik Varma.
\newblock Extreme classification: Tagging on wikipedia, recommendation on
  amazon and advertising on bing.
\newblock In \emph{Proc. The Web Conference}, pp.\  1897, 2018.

\bibitem[Williams et~al.(2018)Williams, Nangia, and Bowman]{MNLI}
Adina Williams, Nikita Nangia, and Samuel Bowman.
\newblock A broad-coverage challenge corpus for sentence understanding through
  inference.
\newblock In \emph{Proc. Association for Computational Linguistics (ACL)},
  2018.

\bibitem[Yin et~al.(2019)Yin, Hay, and Roth]{0SHOT-TC}
Wenpeng Yin, Jamaal Hay, and Dan Roth.
\newblock Benchmarking zero-shot text classification: Datasets, evaluation and
  entailment approach.
\newblock In \emph{Proc. Conf. Empirical Methods Natural Language Processing
  (EMNLP)}, pp.\  3914--3923, 2019.

\bibitem[You et~al.(2019)You, Zhang, Wang, Dai, Mamitsuka, and
  Zhu]{AttentionXML}
Ronghui You, Zihan Zhang, Ziye Wang, Suyang Dai, Hiroshi Mamitsuka, and
  Shanfeng Zhu.
\newblock Attentionxml: Label tree-based attention-aware deep model for
  high-performance extreme multi-label text classification.
\newblock In \emph{Proc. Adv. Neural Information Processing Systems (NeurIPS)},
  2019.

\bibitem[Zeng et~al.(2022)Zeng, Ni, Fang, Li, Zhao, and Song]{WDDC}
Ziqian Zeng, Weimin Ni, Tianqing Fang, Xiang Li, Xinran Zhao, and Yangqiu Song.
\newblock Weakly supervised text classification using supervision signals from
  a language model.
\newblock In \emph{Findings of the Association for Computational Linguistics:
  NAACL}, pp.\  2295--2305, 2022.

\bibitem[Zhang et~al.(2019)Zhang, Lertvittayakumjorn, and Guo]{3CosMul-TC}
Jingqing Zhang, Piyawat Lertvittayakumjorn, and Yike Guo.
\newblock Integrating semantic knowledge to tackle zero-shot text
  classification.
\newblock In \emph{Proc. NAACL - Human Language Technologies}, pp.\
  1031--1040, 2019.

\bibitem[Zhang et~al.(2021)Zhang, Ding, Xu, Liu, and Zhou]{ClassKG}
Lu~Zhang, Jiadong Ding, Yi~Xu, Yingyao Liu, and Shuigeng Zhou.
\newblock Weakly-supervised text classification based on keyword graph.
\newblock In \emph{Proc. Conf. Empirical Methods Natural Language Processing
  (EMNLP)}, pp.\  2803--2813, 2021.

\bibitem[Zhang \& Zhou(2007)Zhang and Zhou]{MLKNN}
M.~Zhang and Z.~Zhou.
\newblock Ml-knn: A lazy learning approach to multi-label learning.
\newblock \emph{Pattern Recognition}, 40\penalty0 (7):\penalty0 2038--2048,
  2007.

\bibitem[Zhang et~al.(2022)Zhang, Yuan, Wang, Bai, and Liu]{LTA}
Yiwen Zhang, Caixia Yuan, Xiaojie Wang, Ziwei Bai, and Yongbin Liu.
\newblock Learn to adapt for generalized zero-shot text classification.
\newblock In \emph{Proc. Association for Computational Linguistics (ACL)}, pp.\
   517--527, 2022.

\bibitem[Zhou et~al.(2020)Zhou, Ma, Long, Xu, Ding, Zhang, Xie, and
  Liu]{HMLTC_Global}
Jie Zhou, Chunping Ma, Dingkun Long, Guangwei Xu, Ning Ding, Haoyu Zhang,
  Pengjun Xie, and G.~Liu.
\newblock Hierarchy-aware global model for hierarchical text classification.
\newblock In \emph{Proc. Association for Computational Linguistics (ACL)},
  2020.

\end{thebibliography}
\bibliographystyle{collas2023_conference}
\newpage
\section*{Appendices}
\addcontentsline{toc}{section}{Appendices}
\renewcommand{\thesubsection}{\Alph{subsection}}

\subsection{Dataset Statistics}
\label{appendix:dataset_statistics}
\begin{table}[h]
    \caption{Dataset Statistics}
    \label{tab:data_statistics}
    \begin{center}
    \resizebox{\textwidth}{!}{
    \begin{tabular}{@{}lccccc@{}} 
    \toprule 
         \textit{Dataset} & \textit{\# of Labels} & \textit{\# of Train} & \textit{\# of Test} & \textit{Ave. Labels per Sample} & \textit{Ave. Samples per Label} \\  \midrule
         \textit{Reuters21578} & 135 & 7694 & 3010 & 1.24 & 70.62 \\
         \textit{StackEx-Philosophy} & 294 & 3983 & 996 & 2.34 & 31.74  \\ \bottomrule
    \end{tabular}}
    \end{center}
\end{table}

\subsection{Evaluation Metrics}
\label{appendix:metrics}
\paragraph{Instance based metrics:} Given $L$ number of labels and $M$ number of samples, let $\mathbf{y}_i = (y_{i1}, \dots y_{iL})$ and $\hat{\mathbf{y}}_i = (\hat{y}_{i1}, ..., \tilde{y}_{iL})$ denote binary vectors that corresponds to ground-truth and predicted labels on sample $i$ respectively. That is:
\begin{equation}
    y_{il} = \begin{cases}
    1,& \text{ if } l \in S_i \\
    0,& \text{ otherwise}
    \end{cases}, 
    \hat{y}_{il} = \begin{cases}
    1,& \text{ if } l \in \hat{S}_i \\
    0,& \text{ otherwise}
    \end{cases}
\end{equation}
Subset accuracy is defined as follows:
\begin{equation}
    \label{eq:ACC}
    \mathrm{ACC} = \frac{1}{M} \sum_{i=1}^M I[\mathbf{y}_i = \hat{\mathbf{y}}_i]
\end{equation}
Hamming accuracy is defined as follows:
\begin{equation}
    \label{eq:HA}
    \mathrm{HA} = \frac{1}{M} \sum_{i=1}^M \frac{1}{L} I[y_{il} = \hat{y}_{il}]
\end{equation}
Example-based F1 score is defined as follows:
\begin{equation}
    \label{eq:ebF1}
    \mathrm{ebF1} = \frac{1}{M} \sum_{i=1}^{M} \dfrac{2 \sum_{l=1}^L y_{il} \hat{y}_{il}}{\sum_{l=1}^L y_{il} + \sum_{l=1}^L \hat{y}_{il}}
\end{equation}

\paragraph{Label based metrics:} Each label $l$ is treated as a separate binary classification problem (each label $l$ has its own confusion matrix of true-positives ($tp_l$), false-positives ($fp_l$), true-negatives ($tn_l$), false-negatives ($fn_l$).)
Micro-averaged F1 score is defined as follows:
\begin{equation}
    \label{eq:miF1}
    \mathrm{miF1} = \dfrac{\sum_{l=1}^L 2 tp_l}{\sum_{l=1}^L 2 tp_l +fp_l +fn_l}
\end{equation}
Macro-averaged F1 score is defined as follows:
\begin{equation}
    \label{eq:maF1}
    \mathrm{maF1} = \dfrac{1}{L} \sum_{l=1}^L \frac{2 tp_l}{2 tp_l +fp_l +fn_l}
\end{equation}

\newpage
\subsection{Sensitivity Study - Noise in Label Frequency Estimates.} 
We conducted an additional sensitivity study on the effect of noise in the input label frequency estimates. We (1) divide the labels into clusters by their observation frequency into $k$ groups, e.g. group 1 higher than 70\% frequency, \dots, group $k$ less than 0.01\% frequency and (2) associate each group member's expected observation frequency to the average observation frequency of its group. The change of resultant performance on annotation-free setting by the number of groups $k$ are presented in Table \ref{tab:sensitivity_noise_frequencies}.

\begin{table}[h]
\caption{Sensitivity study with different numbers of observation frequency-based label groups (k) in annotation-free setting. All results are calculated over 10 random initialization on the original train-test data splits.}
\label{tab:sensitivity_noise_frequencies}
\begin{center}
\resizebox{\textwidth}{!}{
\begin{tabular}{@{}llcccccccccc@{}}
\toprule
                      &        & \multicolumn{5}{c}{\textit{Reuters21578}}           & \multicolumn{5}{c}{\textit{StackEx-Philosophy}}     \\ 
                      &        & ACC    & HA     & ebF1   & miF1   & maF1   & ACC    & HA     & ebF1   & miF1   & maF1   \\ \midrule
0Shot-MLTC &  & 0.0834 & 0.9799 & 0.2778 & 0.2981 & 0.1844 & 0.0010 & 0.8802 & 0.0924 & 0.0665 & 0.1528 \\ \midrule
BNCL & k = 2  & 0.0950 & 0.9903 & 0.1654 & 0.2401 & 0.1663 & 0.0226 & 0.9900 & 0.1899 & 0.2218 & 0.2290 \\
     & k= 4   & 0.2753 & 0.9918 & 0.3926 & 0.4612 & 0.1740 & 0.0355 & 0.9901 & 0.2078 & 0.2390 & 0.2294 \\
     & k = 6  & 0.2758 & 0.9918 & 0.3936 & 0.4621 & 0.1741 & 0.0358 & 0.9902 & 0.2084 & 0.2392 & 0.2302 \\
     & k = 8  & 0.2765 & 0.9918 & 0.3945 & 0.4631 & 0.1740 & 0.0363 & 0.9902 & 0.2096 & 0.2399 & 0.2298 \\
     & k = 10 & 0.2770 & 0.9918 & 0.3950 & 0.4637 & 0.1743 & 0.0370 & 0.9902 & 0.2108 & 0.2415 & 0.2303 \\ \bottomrule
\end{tabular}
}
\end{center}
\end{table}

Based on this sensitivity study, we conclude that if the domain-supervisor can reasonably group the labels into four by their observation frquency, e.g. almost unseen, uncommon, standard, common, the approximate identification leads to a dramatic improvement compared to the baseline performance on the datasets under interest. 

\end{document}